\documentclass[letterpaper, 10 pt, conference]{ieeeconf}  % Comment this line out if you need a4paper

\IEEEoverridecommandlockouts                              % This command is only needed if 
\usepackage{amsmath}
\usepackage{amssymb}
\usepackage{graphicx}
\usepackage{xcolor}
\usepackage{url}

\title{\LARGE \bf
Recon2Servo: Robotic Ultrasound \\Visual Servoing via Learned Image-to-Motion Inference
}

\author{
Yameng Zhang$^{1,2,3}$,
Pei Liu$^{1,4}$,
Dianye Huang$^{1}$,
Yizhao Qian$^{5}$, \\
Zhongyu Chen$^{6}$,
Xiangyu Chu$^{2,3}$,
K. W. Samuel Au$^{2,3}$,
and Zhongliang Jiang$^{1,*}$
\thanks{$^{*}$Corresponding author: Zhongliang Jiang
(email: zljiang@hku.hk).}
\thanks{$^{1}$Department of Mechanical Engineering, The University of Hong Kong, Hong Kong SAR.
$^{2}$Multi-scale Medical Robotics Center, AIR@InnoHK, Hong Kong SAR.
$^{3}$Department of Mechanical and Automation Engineering, The Chinese University of Hong Kong, Hong Kong SAR.
$^{4}$School of Computation, Information and Technology, Technical University of Munich, Germany.
$^{5}$Department of Electronic Engineering, The Chinese University of Hong Kong, Hong Kong SAR.
$^{6}$School of Optometry, Hong Kong Polytechnic University, Hong Kong SAR.}
}

\begin{document}
\maketitle
\thispagestyle{empty}
\pagestyle{empty}

%%%%%%%%%%%%%%%%%%%%%%%%%%%%%%%%%%%%%%%%%%%%%%%%%%%%%%%%%%%%%%%%%%%%%%%%%%%%%%%%
\begin{figure*}[!t]
    \centering
    \includegraphics[width=0.95\textwidth]{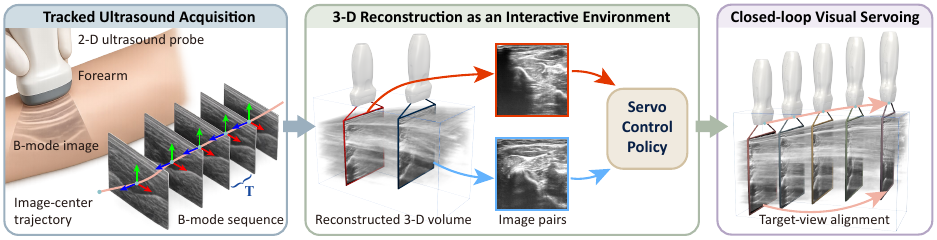}
    \caption{Overview of the proposed ultrasound servoing framework. Tracked B-mode images are reconstructed into a 3-D volume, which serves as an interactive simulation environment for training and adapting the probe control policy. The resulting policy is deployed using real-time ultrasound image feedback to iteratively guide the probe toward a target view.}
    \label{fig:recon2servo_overview}
\end{figure*}

%%%%%%%%%%%%%%%%%%%%%%%%%%%%%%%%%%%%%%%%%%%%%%%%%%%%%%%%%%%%%%%%%%%%%%%%%%%%%%%%
\begin{abstract}
Ultrasound visual servoing is essential for autonomous robotic ultrasound, yet 6-DoF probe control from 2D B-mode images remains challenging due to limited and ambiguous out-of-plane motion cues. Existing methods typically rely on anatomical priors or handcrafted visual features, limiting their generalizability across imaging targets. Inspired by trackerless 3D ultrasound reconstruction, we propose \textbf{Recon2Servo}, a visual servoing framework that learns image-to-motion inference directly from B-mode images for 6-DoF probe control. A DINOv3 encoder with low-rank adaptation and a bidirectional relation module estimate the relative probe pose between current and target images to guide iterative closed-loop target-view alignment. The framework combines supervised relative-pose learning, reconstruction-guided closed-loop adaptation, and bounded residual pose correction to improve motion inference during servoing. Evaluations on a public dataset and an in-house dataset collected from 12 healthy volunteers using different ultrasound systems demonstrate its effectiveness in reconstructed-volume servoing. Additional real-robot demonstrations of target-view alignment and dynamic tracking on a human forearm are provided in the supplementary video: \url{https://youtu.be/qwsOdI-GMYk}.
\end{abstract}

%%%%%%%%%%%%%%%%%%%%%%%%%%%%%%%%%%%%%%%%%%%%%%%%%%%%%%%%%%%%%%%%%%%%%%%%%%%%%%%%
\section{INTRODUCTION}
\par
Ultrasound visual servoing is a critical technology for robotic ultrasound, as it enables the robot to directly exploit real-time imaging feedback to regulate probe motion according to the observed anatomy, thereby supporting autonomous probe navigation, image stabilization, and target tracking~\cite{jiang2023robotic}. However, conventional medical ultrasound typically provides cross-sectional 2D images. While in-plane motion can be relatively easily estimated from abundant visual features, out-of-plane probe motion is long-standing challenging to infer because it causes anatomical structures to disappear from or newly enter the imaging plane, providing limited and often ambiguous motion cues.

\par
In a pioneering study, Abolmaesumi~\emph{et al.} proposed a shared-autonomy strategy combining human expertise with ultrasound visual servoing~\cite{abolmaesumi2001user}. The in-plane 3-DoF probe motion was automatically controlled to center the carotid artery in cross-sectional ultrasound images, while the remaining three out-of-plane DoFs were teleoperated by a human expert. Recently, Qian~\emph{et al.} employed optical flow as a feature for high-frequency 3-DoF translational tracking~\cite{qian2025breaking}. Extending ultrasound visual servoing to fully control the 6-DoF probe motion, however, requires servoing features that encode both in-plane and out-of-plane motion. In 2006, Bachta~\emph{et al.} developed a servoing method to drive the probe from an arbitrary initial pose toward a desired ultrasound image~\cite{bachta2006towards}. The features were derived from object boundaries extracted from ultrasound images, with the target contour recovered using polynomial regression tailored to simplified egg-shaped objects.

\par
To enable generic visual servoing without prior knowledge of object geometry, Krupa~\emph{et al.} exploited ultrasound speckle to estimate both in-plane and out-of-plane probe motion~\cite{krupa2007real,krupa2009real}. Speckle patterns exhibit spatial coherence across consecutive ultrasound images, providing intrinsic visual cues for estimating probe motion. However, speckle also degrades image quality and is therefore commonly suppressed by despeckling and image-enhancement techniques in modern ultrasound systems~\cite{li2025speckle2self}, limiting its reliability and generalizability as a visual servoing feature. Beyond speckle-based approaches, alternative image representations have been explored to construct features for 6-DoF visual servoing. Mebarki~\emph{et al.} employed image moments to characterize the spatial distribution of image content~\cite{mebarki20102, mebarki2009modeling}, while Nadeau~\emph{et al.} directly exploited image intensities as servoing features~\cite{nadeau2013intensity}. Moving toward less handcrafted feature representations, Duflot~\emph{et al.} first introduced subsampled shearlet coefficients as visual features for the control law~\cite{duflot2019wavelet}. By providing a redundant yet noise-robust representation of ultrasound image content in a selected frequency domain, the shearlet coefficients enabled accurate, repeatable, and robust visual servoing, as demonstrated in both simulation and phantom experiments.

\par
% More recently, task-specific anatomical cues have been incorporated into ultrasound visual servoing. Ma~\emph{et al.} employed anatomical template matching for local probe navigation toward standardized lung imaging planes~\cite{ma2024guiding}, while Qian~\emph{et al.} combined image symmetry with biomechanics-based force feedback to improve probe alignment~\cite{qian2026symmetry}. While effective for their targeted applications, these approaches rely on anatomy- or task-specific priors, which may limit their generalizability to different anatomical structures and imaging scenarios. 

With the rapid advances in deep learning, recent studies have increasingly explored reinforcement and imitation learning for standard-plane navigation. Li~\emph{et al.} formulated 6-DoF standard-plane navigation as a reinforcement learning problem with a handcrafted reward function~\cite{li2021autonomous}. To reduce the reliance on manually designed rewards, Jiang~\emph{et al.} employed inverse reinforcement learning to infer image-based rewards from a small number of expert demonstrations~\cite{jiang2024intelligent}. H.~Jiang~\emph{et al.} further adopted imitation learning to directly learn probe navigation from demonstrations, demonstrating its effectiveness in standardized carotid ultrasound scanning workflows~\cite{jiang2025towards}. However, these approaches are typically trained for predefined target planes or scanning tasks and therefore cannot readily generalize to arbitrarily specified imaging planes, due to the lack of a reliable mapping between ultrasound image variations and the corresponding spatial probe motion.

\par
To address this challenge, we draw inspiration from trackerless freehand 3D ultrasound reconstruction, which has attracted increasing attention in recent years. At its core, trackerless reconstruction aims to infer the spatial motion of the ultrasound probe directly from image sequences, as demonstrated by early inter-frame pose regression methods~\cite{prevost20183d}. Recent studies have further improved motion estimation by modeling long-term sequence dependencies~\cite{li2023long}, incorporating physics-guided speckle correlation~\cite{dou2024sensorless}, learning fine-grained spatiotemporal representations with inertial alignment~\cite{yan2024fine}, and separately encoding local motion and global anatomical context, as in DualTrack~\cite{wilson2025dualtrack}. Collectively, these advances demonstrate that spatial probe motion can be inferred from ultrasound image sequences by jointly exploiting local image variations and broader anatomical context. Inspired by these advances, we explore whether the image-to-motion inference capability developed for trackerless reconstruction can be leveraged to establish a generalizable spatial representation for ultrasound visual servoing.

\par
In this study, we propose \textbf{Recon2Servo}, a novel ultrasound visual servoing framework that directly learns 6-DoF probe motion from B-mode ultrasound images. To the best of our knowledge, Recon2Servo is the first approach to achieve 6-DoF ultrasound visual servoing through learned image-to-motion inference, without relying on sophisticated handcrafted visual features, whose design is particularly challenging in ultrasound due to substantial inter-operator variability and inherent imaging noise. Specifically, a DINOv3 encoder~\cite{simeoni2025dinov3}, adapted using low-rank adaptation (LoRA)~\cite{hu2021lora}, extracts multi-level visual features that are fed into a bidirectional relation module with dedicated translation and rotation motion tokens to infer the relative 6-DoF pose between the current and target images. The predicted relative pose is directly converted into probe motion through a fixed-gain controller. To improve robustness to states encountered during closed-loop servoing, we further introduce a dataset aggregation (DAgger)-style adaptation strategy~\cite{ross2011reduction}, which updates the pose estimator using labeled states visited during tracked reconstruction trajectories. Finally, a bounded residual head leverages frozen relational features and the mean pose estimate to further refine the motion prediction. To evaluate the proposed method, we assess relative-pose estimation accuracy and closed-loop target-view convergence using the public TUS-REC2024 dataset~\cite{li2025tus} and an in-house ultrasound dataset collected from 12 healthy volunteers. These evaluations examine the effectiveness of the learned image-to-motion representation and its adaptation across ultrasound acquisition domains.

\section{METHODS}
\label{sec:method}
\par
Drawing inspiration from learning-based trackerless 3D ultrasound reconstruction, Recon2Servo employs a motion-aware visual perception module to infer the relative 6-DoF probe motion between a pair of B-mode ultrasound images. To generate image pairs covering diverse relative motions across all six DoFs, we first reconstruct each tracked ultrasound scan into a 3D volume using the recorded probe poses and then reslice the volume at different poses to synthesize paired B-mode images with known relative transformations. The architecture for estimating the 6-DoF relative pose between two images is illustrated in Fig.~\ref{fig:recon2servo_overview}. During servoing, the current and target images are provided to the motion estimator, whose predicted relative pose guides the probe toward the target view. After each probe-pose update, a new current image is acquired, enabling iterative closed-loop alignment with the target image. The corresponding tracked poses provide supervision for the states encountered throughout this process.

\par
The overall pipeline consists of three stages. Stage 1 trains a relative-pose estimator on paired ultrasound images with known 6-DoF transformations. Stage 2 adapts the estimator using labeled states encountered during closed-loop rollouts within reconstructed ultrasound volumes, improving robustness to the state distribution induced by iterative servoing. Stage 3 learns a bounded residual correction to further refine the pose predictions. The resulting relative-pose estimate is converted into probe motion through a fixed-gain control law. During deployment, live ultrasound images provide continuous visual feedback for closed-loop probe control.

% Recon2Servo uses recoded tracked ultrasound dataset to learn image-based probe control toward a desired view. As illustrated in Fig.~\ref{fig:recon2servo_overview}, tracked B-mode images and their spatial poses are used to reconstruct a 3-D volume that serves as an interactive servoing environment. It renders current and target views for the control policy and returns a new current view after each probe-pose update, enabling iterative target-view alignment. The known poses provide supervision for the visited states.

% We develop the controller through three learning stages. Stage 1 trains a relative-pose estimator on recorded image pairs. Stage 2 adapts it using labeled states collected during closed-loop rollouts in reconstructed volumes. Stage 3 learns a bounded residual correction to refine its predictions. The resulting pose estimate is converted into probe motion through a fixed-gain control law; during deployment, live ultrasound images provide the feedback.

%%%%%%%%%%%%%%
\subsection{Supervised Inter-Image Pose Learning}
% \subsection{Stage 1: Supervised Image-to-Motion Learning}
% \par
% Stage 1 learns to estimate relative probe pose from tracked image pairs.
% We define the pose labels before introducing the estimator and its training objectives.

\subsubsection{Relative Pose Representation}
\par
For each current-target image pair $(I_i,I_j)$, the tracked poses
$\mathbf{T}_i,\mathbf{T}_j\in \mathrm{SE}(3)$ map the calibrated image-center frames to a common reference frame. These poses define the relative transformation used to supervise the neural estimator, which we express as a six-dimensional twist:
\begin{equation}
 \mathbf{G}_{ij}=\mathbf{T}_i^{-1}\mathbf{T}_j,\quad
 \boldsymbol{\xi}_{ij}=\operatorname{Log}(\mathbf{G}_{ij})^\vee.
 \label{eq:relative_pose}
\end{equation}
where $\boldsymbol{\xi}_{ij}\in\mathbb{R}^{6}$ contains translational and rotational
twist components. The operator $\vee$ maps a Lie-algebra matrix to its twist
vector, and $\wedge$ denotes the inverse mapping. To balance the contribution
of these components to the training loss, we normalize each by its training-set
standard deviation:
\begin{equation}
 \bar{\boldsymbol{\xi}}_{ij}=\mathbf{S}^{-1}\boldsymbol{\xi}_{ij}.
 \label{eq:normalized_pose}
\end{equation}
where $\mathbf{S}=\operatorname{diag}(s_1,\ldots,s_6)$ stores these standard deviations
and remains fixed during adaptation and inference. The resulting normalized
pose $\bar{\boldsymbol{\xi}}_{ij}$ serves as the training target for the network,
which takes only $(I_i,I_j)$ as input and produces a deterministic estimate
$\boldsymbol{\mu}_{ij}$. We refer to this output as the mean prediction
throughout the following stages.

\begin{figure*}[t]
    \centering
    \includegraphics[width=0.95\textwidth]{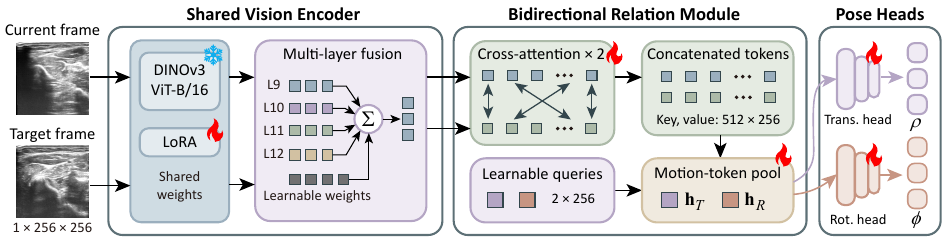}
    \caption{Relative-pose estimation network. A shared LoRA-adapted DINOv3 encoder and bidirectional relation module map current and target images to translation and rotation predictions. Snowflakes and flames denote frozen and trainable components, respectively.}
    \label{fig:stage1}
\end{figure*}

\subsubsection{Network Architecture}

As shown in Fig.~\ref{fig:stage1}, the pose estimator comprises a shared image
encoder, a bidirectional relation module, and separate translation and rotation
heads. Together, these components map the current and target images to the
normalized relative-pose estimate $\boldsymbol{\mu}_{ij}$.
Each grayscale input is replicated across
three channels and normalized using the encoder's ImageNet statistics. Both
images pass through the same visual foundation model DINOv3 ViT-B/16 encoder~\cite{simeoni2025dinov3}. Its original weights remain frozen, while rank-8 LoRA~\cite{hu2021lora} adapts the query and value projections of the final two blocks. This preserves the pretrained representation while allowing ultrasound-specific adjustment through a small set of parameters. Patch tokens from the final four layers are fused using softmax-normalized layer weights:
\begin{equation}
 \mathbf{F}(I)=\sum_{\ell\in\mathcal{K}}\alpha_\ell \mathbf{F}_\ell(I),\quad
 \alpha_\ell=\frac{\exp(a_\ell)}{\sum_{m\in\mathcal{K}}\exp(a_m)}.
 \label{eq:feature_fusion}
\end{equation}
where $\mathcal{K}$ indexes these layers and $a_\ell$ are trainable scalars shared across images. The softmax fusion illustrated in Fig.~\ref{fig:stage1}
lets motion supervision select the contribution of each feature layer.
For an input $256\times256$, the fused representation contains 256 spatial tokens of dimension 768.

\par
The fused tokens are projected to 256 dimensions and augmented with 2-D position
and current/target type embeddings. Two bidirectional cross-attention blocks,
each with eight heads, update both image streams from their paired inputs.
This allows motion inference to depend on spatial evidence from both views.
The updated streams are concatenated into 512 relation tokens, which serve as
keys and values for two learned motion queries. Pooling produces translation and
rotation tokens $\mathbf{h}_T,\mathbf{h}_R\in\mathbb{R}^{256}$. Two pose heads
predict the normalized translation and rotation components ($\rho$ and $\phi$
in Fig.~\ref{fig:stage1}):
\begin{equation}
 \boldsymbol{\mu}_{ij}
 =\begin{bmatrix}\hat{\boldsymbol{\rho}}_{ij}\\\hat{\boldsymbol{\phi}}_{ij}\end{bmatrix}
 =\begin{bmatrix}p_T(\mathbf{h}_T)\\p_R(\mathbf{h}_R)\end{bmatrix}.
 \label{eq:mean_prediction}
\end{equation}
Each head is a $256\!\rightarrow\!256\!\rightarrow\!128\!\rightarrow\!3$ MLP
with GELU and dropout.

\begin{figure*}[t]
    \centering
    \includegraphics[width=0.95\textwidth]{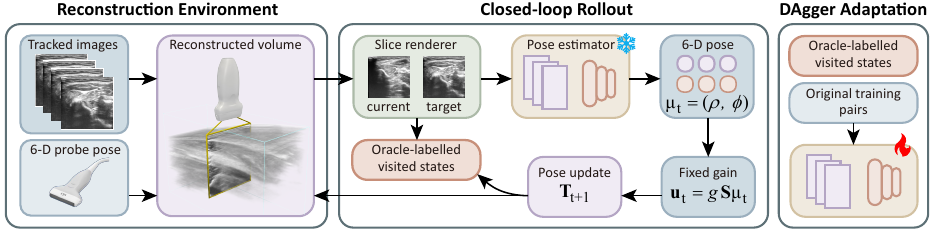}
    \caption{Reconstruction-guided closed-loop adaptation. Tracked images and calibrated poses define the reconstruction environment. A frozen pose estimator drives fixed-gain rollouts, and the renderer returns images at the updated poses. The known current and target poses provide oracle labels for visited states, which are combined with recorded training pairs to update the pose estimator.}
    \label{fig:stage2}
\end{figure*}

% \begin{figure*}[t]
%     \centering
%     \includegraphics[width=0.9\textwidth]{figure/stage3.pdf}
%     \caption{Bounded residual pose correction. The adapted encoder, relation module, and pose heads are frozen. The two motion tokens and normalized mean estimate are concatenated and passed to a trainable residual MLP. Its output is bounded in physical twist coordinates and added to the mean.}
%     \label{fig:stage3}
% \end{figure*}

\begin{figure}[t]
    \centering
    \includegraphics[width=0.85\columnwidth]{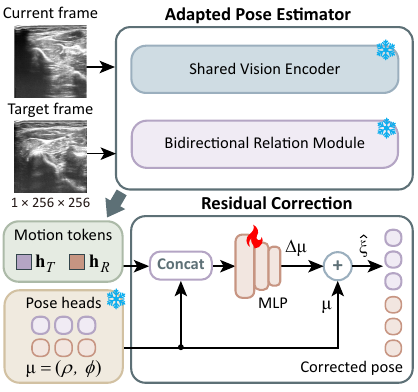}
    \caption{Bounded residual pose correction. The adapted encoder, relation module, and pose heads are frozen. The two motion tokens and normalized mean estimate are concatenated and passed to a trainable residual MLP. Its output is bounded in physical twist coordinates and added to the mean.}
    \label{fig:stage3}
\end{figure}

%%%%%%%%%%%%%%%%%%%%%%%%%%%%%%%%%%%%%%
\subsubsection{Data Augmentation and Training Objectives}
To improve robustness to ultrasound appearance variations, we augment image
intensity, sharpness, and noise while preserving the field of view and pose labels. Intensity augmentation adjusts brightness, contrast, gamma, and dynamic range, and introduces depth-dependent attenuation. Sharpness is varied through Gaussian blur and downsampling followed by upsampling. Additive and multiplicative Gaussian noise provide further variation in image quality.

\par
For each training pair $(I_i,I_j)$, we generate two augmented variants,
$(I_i^a,I_j^a)$ and $(I_i^b,I_j^b)$, with augmentation parameters sampled
independently for each image and variant. The shared network processes each
variant in both forward and reverse order, producing four normalized pose
estimates $\boldsymbol{\mu}_d^v$, where $d\in\{ij,ji\}$ denotes direction and
$v\in\{a,b\}$ denotes the appearance variant. These predictions support \textbf{three complementary objectives}: supervised pose accuracy, inverse consistency, and consistency across appearance variants. All three objectives use the component-averaged Smooth-$L_1$:
\begin{equation}
 \ell_\beta(\mathbf{e})=\frac{1}{6}\sum_{k=1}^{6}
 \begin{cases}
 e_k^2/(2\beta),& |e_k|<\beta,\\
 |e_k|-\beta/2,& \text{otherwise},
 \end{cases}
 \label{eq:smooth_l1}
\end{equation}
with $\beta=0.5$, where $\mathbf{e}$ is a 6D error vector.
For pose supervision, we compare each prediction with its normalized
ground-truth pose and average over the four predictions:
\begin{equation}
 \mathcal{L}_{\mathrm{pose}}
 =\frac14\sum_{v\in\{a,b\}}\sum_{d\in\{ij,ji\}}
 \ell_\beta(\boldsymbol{\mu}_{d}^{v}-\bar{\boldsymbol{\xi}}_{d}).
 \label{eq:pose_loss}
\end{equation}
To enforce inverse consistency, we convert each prediction to a rigid
transformation, $\hat{\mathbf{G}}_d^v=\operatorname{Exp}((\mathbf{S}\boldsymbol{\mu}_d^v)^\wedge)$,
and penalize deviations of the forward-reverse composition from the identity:
\begin{equation}
 \mathcal{L}_{\mathrm{bi}}
 =\frac12\sum_{v\in\{a,b\}}
 \ell_\beta\!\left(\mathbf{S}^{-1}\operatorname{Log}
 (\hat{\mathbf{G}}_{ij}^{v}\hat{\mathbf{G}}_{ji}^{v})^\vee\right).
 \label{eq:bidirectional_loss}
\end{equation}
Because augmentation preserves the underlying pose, the two appearance variants
should yield the same estimate for a given direction. We encourage this through
\begin{equation}
 \mathcal{L}_{\mathrm{sty}}
 =\frac12\sum_{d\in\{ij,ji\}}
 \ell_\beta(\boldsymbol{\mu}_{d}^{a}-\boldsymbol{\mu}_{d}^{b}).
 \label{eq:style_loss}
\end{equation}
The combined training loss is
\begin{equation}
 \mathcal{L}_{\mathrm{mean}}
 =\mathcal{L}_{\mathrm{pose}}+0.1\mathcal{L}_{\mathrm{bi}}
   +0.1\mathcal{L}_{\mathrm{sty}}.
 \label{eq:mean_loss}
\end{equation}
Stage 1 minimizes the expected loss over image pairs sampled from the training
dataset $\mathcal{D}$:
\begin{equation}
 \mathcal{J}_{\mathrm{mean}}(\theta;\mathcal{D})
 =\mathbb{E}_{\mathcal{D}}[\mathcal{L}_{\mathrm{mean}}].
 \label{eq:mean_objective}
\end{equation}
Here, $\theta$ contains the trainable LoRA, feature-fusion, relation-module,
and pose-head parameters, and the expectation follows the training sampler.

%%%%%%%%%%%%%%%%%%%%%%%%%%%%%%%%
\subsection{Reconstruction-Guided Closed-Loop Pose Adaptation}
\par
During closed-loop servoing, prediction errors affect subsequent probe poses and can lead the controller beyond the recorded scan trajectory. The resulting views, together with interpolation effects introduced by volume reslicing, differ from the image pairs used in Stage 1. Stage 2 addresses this distribution shift by adapting the neural estimator to labeled states encountered during its own rollouts in the reconstructed 3D visual environment.

\par
The reconstructed volume is denoted as $V$. Using the calibrated image geometry, a slice renderer $\mathcal{R}$ generates current and target images at pose matrices $\mathbf{T}_t,\mathbf{T}_\star\in \mathrm{SE}(3)$:
\begin{equation}
 I_t=\mathcal{R}(V,\mathbf{T}_t),\quad
 I_\star=\mathcal{R}(V,\mathbf{T}_\star).
 \label{eq:volume_rendering}
\end{equation}
The current view is updated after each probe-pose update, while the target view remains fixed. At every visited pose, the known relative transform $\mathbf{T}_t^{-1}\mathbf{T}_\star$ provides the normalized motion label. This allows the environment to label states beyond the recorded trajectory without requiring expert demonstrations.

\par
As shown in Fig.~\ref{fig:stage2}, we hold the Stage-1 neural estimator fixed and run the fixed-gain controller from sampled initial poses in training volumes. The visited image pairs and their motion labels form:
\begin{equation}
 \mathcal{D}_{\mathrm{visit}}
 =\{(I_t,I_\star,\bar{\boldsymbol{\xi}}_{t\star})\},
 \label{eq:visited_dataset}
\end{equation}
where only pairs with sufficient volume support and displacement within the training range are retained. Following the dataset aggregation principle of DAgger~\cite{ross2011reduction}, we combine these visited pairs with recorded pairs $\mathcal{D}_{\mathrm{rec}}$ and fine-tune the pose estimator using the Stage-1 objective:
\begin{equation}
 \min_\theta\;
 \eta\mathcal{J}_{\mathrm{mean}}(\theta;\mathcal{D}_{\mathrm{rec}})
 +(1-\eta)\mathcal{J}_{\mathrm{mean}}(\theta;\mathcal{D}_{\mathrm{visit}}).
 \label{eq:dagger_adaptation}
\end{equation}
where $\eta$ denotes the fraction of recorded-pair samples and is set to $0.8$ for training on the public dataset. Mixing the two data sources preserves supervision from recorded scans while adapting the pose estimator to states encountered during closed-loop servoing. We perform a single collection-and-update round following a DAgger-style adaptation scheme, during which the LoRA adapters, feature-fusion weights, relation module, and pose heads are jointly optimized.

% where $\eta$ is the recorded-pair sampling fraction, set to $0.8$ for the training on public dataset. Mixing the two sources preserves supervision on recorded scans while adapting the pose estimator to the states encountered during servoing. We perform a single collection-and-update round, yielding a DAgger-style adaptation process, which optimizes LoRA adapters, feature-fusion weights, relation module, and pose heads.

%%%%%%%%%%%%%%%%%%%%%%%%%%%%%%%%%
\subsection{Bounded Residual Pose Correction}
\par
This stage refines the errors remaining after closed-loop adaptation. As shown in Fig.~\ref{fig:stage3}, a residual head receives the two motion tokens and the adapted mean prediction. The mean provides an explicit estimate of displacement, while the motion tokens retain image-pair evidence beyond 6D output. Conditioning on both enables the head to learn corrections that depend on the estimated motion and its supporting relational features. In this stage, the entire adapted pose estimator, including its LoRA adapters, is frozen. Only the residual head is optimized, preserving the base mapping while learning a bounded correction to its output.

\par
Let $r_\phi:\mathbb{R}^{518}\rightarrow\mathbb{R}^{6}$ be an MLP with input
layer normalization and two 256-unit hidden layers. The correction head first
forms its conditioning vector by concatenating the two motion tokens with the
mean prediction:
\begin{equation}
 \mathbf{z}_{ij}=\operatorname{concat}(\mathbf{h}_T,\mathbf{h}_R,
 \boldsymbol{\mu}_{ij}).
 \label{eq:correction_input}
\end{equation}
The output is bounded and returned to normalized coordinates:
\begin{equation}
 \Delta\boldsymbol{\mu}_{ij}=\mathbf{S}^{-1}\Pi\!\left(\mathbf{S}r_\phi(\mathbf{z}_{ij})\right).
 \label{eq:bounded_correction}
\end{equation}
Where $\Pi$ projects the translational and rotational twist blocks onto Euclidean balls of radii 3~mm and $3^\circ$, respectively. Since these bounds are physical and independent of the normalization scales, the correction is mapped to physical units by $\mathbf{S}$ before clipping and back by $\mathbf{S}^{-1}$ afterward. Adding the correction to the mean prediction gives the refined pose in physical units:
\begin{equation}
 \hat{\boldsymbol{\xi}}_{ij}=\mathbf{S}(\boldsymbol{\mu}_{ij}
 +\Delta\boldsymbol{\mu}_{ij}).
 \label{eq:residual_prediction}
\end{equation}
The final MLP layer is initialized to zero, so the initial corrected prediction equals the adapted mean. The head reuses $\mathcal{D}_{\mathrm{visit}}$ from Stage 2 and minimizes
\begin{equation}
 \mathcal{J}_{\mathrm{corr}}(\phi)
 =\mathbb{E}_{\mathcal{D}_{\mathrm{visit}}}\!\left[
 \ell_\beta\!\left(\Delta\boldsymbol{\mu}_{ij}
 -(\bar{\boldsymbol{\xi}}_{ij}-\boldsymbol{\mu}_{ij})\right)\right].
 \label{eq:residual_objective}
\end{equation}

\subsection{Closed-Loop Ultrasound Visual Servoing}
\par
At each step $t$, the current image and fixed target image are processed to obtain the corrected relative twist $\hat{\boldsymbol{\xi}}_{t\star}$. A fixed-gain controller converts this estimate into a local probe motion:
\begin{equation}
 \mathbf{u}_t=\mathcal{C}(g\hat{\boldsymbol{\xi}}_{t\star}),\quad
 \mathbf{T}_{t+1}=\mathbf{T}_t\operatorname{Exp}(\mathbf{u}_t^\wedge),
 \label{eq:servo_control}
\end{equation}
where $g=0.5$ and $\mathcal{C}$ limits the translation and rotation of the commanded increment. After each motion, a new current image is obtained and re-estimated against the target: rendered from the reconstructed volume at $\mathbf{T}_{t+1}$ in virtual servoing, or acquired by the probe in robotic execution, with calibrated geometry mapping the local image-center increment into the robot command frame. The loop thus closes through successive image pairs.

%%%%%%%%%%%%%%%%%%%%%%%%%%%%%%%%%%%%%
\section{EXPERIMENTAL VALIDATION}
\label{sec:experiments}
\par
We evaluate Recon2Servo on the public TUS-REC2024 dataset and an in-house dataset acquired on volunteers arm using different ultrasound systems. The experiments systematically assess the performance gains from closed-loop adaptation and bounded residual pose correction.

\subsection{Datasets and Preprocessing}
\textit{Public ultrasound dataset:}
We use the tracked freehand ultrasound sequences from TUS-REC2024\footnote{\url{https://zenodo.org/records/11178509}} for supervised pose learning and volume-based servoing. We split the dataset by subject into 40 training, 5 validation, and 5 test subjects, comprising 480, 60, and 60 sequences, respectively. From each frame, we extract a horizontally centered $30\times30$~mm region of interest (ROI) and resize it to $256\times256$ pixels, yielding a pixel spacing of $0.117$~mm. We update the image-center poses and rendering geometry accordingly to preserve the physical meaning of the motion labels. Training pairs are sampled with a maximum frame gap of 80 and a relative translation below 15~mm, and their pose statistics define the normalization matrix $\mathbf{S}$. For virtual servoing, we reconstruct volumes over the original scan coverage and render observations using the ROI geometry.

\begin{figure}[t]
    \centering
    \includegraphics[width=0.9\linewidth]{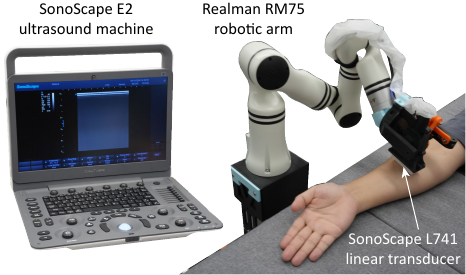}
    \caption{Robotic ultrasound acquisition setup for the in-house dataset. A SonoScape L741 linear transducer connected to a SonoScape E2 system is mounted on a Realman RM75 robotic arm for automated forearm scanning.}
    \label{fig:acquisition_setup}
\end{figure}

\textit{In-house ultrasound dataset\footnote{Data will be released upon paper acceptance.}:}
We acquired forearm ultrasound scans from 12 healthy volunteers using a portable SonoScape E2 system and an L741 linear transducer mounted on a Realman RM75 robotic arm (Fig.~\ref{fig:acquisition_setup}). B-mode images were acquired at a frequency setting of 9.5-12.2~MHz, a depth of 50~mm, and a nominal frame rate of 40~fps. For each participant, the robot scanned both forearms along C-, L-, and S-shaped trajectories in both distal-to-proximal and proximal-to-distal directions, yielding 12 sequences per participant and 144 in total. The data were split by participant into training, validation, and test sets of 8, 2, and 2 participants, respectively. Preprocessing uses the same $30\times30$~mm ROI size and $256\times256$ input resolution as the public dataset. Recorded robot poses are transformed into the calibrated image-center coordinate system, and the original images are compounded into volumes with a voxel spacing of $0.5$~mm. Virtual observations are rendered from these volumes using the ROI geometry. Image pairs are sampled with a maximum frame gap of 120 and a maximum relative translation of 15~mm.

\begin{figure}[t]
    \centering
    \includegraphics[width=0.8\linewidth]{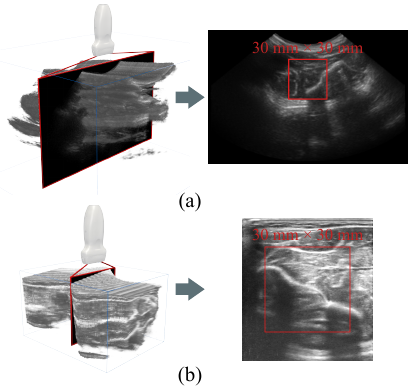}
    \caption{Reconstructed ultrasound volumes and image regions used for servoing in (a) the public TUS-REC2024 dataset and (b) the in-house dataset. Left: reconstructed volumes with the imaging planes outlined in red. Right: corresponding B-mode views, with red boxes indicating the $30\times30$~mm ROIs resized to $256\times256$ pixels for model input.}
    \label{fig:dataset_comparison}
\end{figure}

Fig.~\ref{fig:dataset_comparison} illustrates the reconstructed volumes and representative B-mode views from the two datasets. The curvilinear and linear acquisitions differ in field of view and image appearance; both are processed using a common $30\times30$~mm ROI and $256\times256$ input resolution to standardize the physical extent and sampling of the model inputs.

\begin{figure*}[t]
    \centering
    \includegraphics[width=0.85\textwidth]{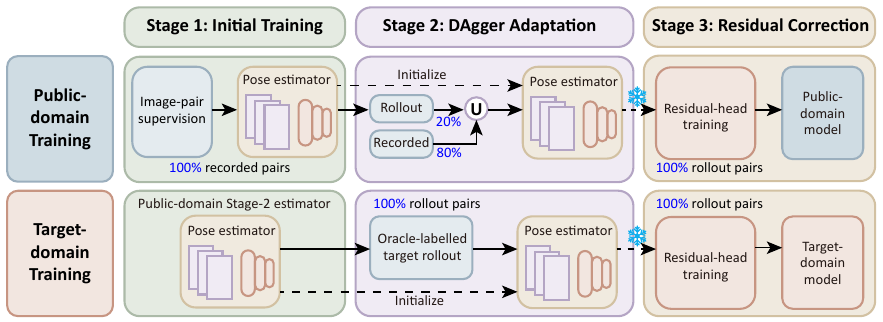}
    \caption{Dataset-specific training protocols. The public-dataset estimator is trained on recorded pairs, adapted using 80\% recorded and 20\% rollout pairs, and refined using rollout pairs alone. For the in-house dataset, the public-dataset Stage-2 estimator initializes adaptation using labeled in-house rollout pairs, followed by residual-head training on the same rollout data.}
    \label{fig:data}
\end{figure*}

%%%%%%%%%%%%%%%%%%%%%%%%%%%%%%%%%%%%%
\subsection{Training Protocols}
\label{sec:training_protocols}

Fig~\ref{fig:data} summarizes the data usage and parameter initialization for the two datasets. The upper row shows the three-stage public-dataset protocol. Stage~1 trains the pose estimator entirely on recorded image pairs. The resulting estimator generates closed-loop rollouts in training volumes, and Stage~2 updates it using a mixture of 80\% recorded pairs and 20\% oracle-labeled rollout pairs. Stage~3 freezes the adapted estimator and trains the residual head entirely on the same rollout pairs. Reusing the collected states ensures that the residual stage does not benefit from an additional round of data collection.

\par
The lower row shows the in-house protocol. The public-dataset Stage-2 estimator initializes the in-house pose estimator and serves as the behavior policy for collecting labeled states in the in-house training volumes. Adaptation uses these rollout pairs alone, without recorded-pair replay. The adapted estimator is then frozen, and a residual head is trained on the same in-house rollout dataset. Thus, both protocols use reconstruction-derived supervision to improve behavior on controller-visited states.

\par
One rollout-collection round is used for each protocol. Public-dataset collection retains at most 20 sampled states per episode; in-house collection retains at most 50. Collection stops when the reaching criterion is met, and retained pairs must have at least 90\% slice coverage and relative translation within 15~mm. The in-house models initialized from the public estimator retain its pose-normalization scales throughout adaptation and inference.

All models are trained on two NVIDIA GeForce RTX 5090 GPUs using AdamW, a batch size of 32, 500 steps per epoch, mixed precision, and gradient clipping at a norm of 1.0.
% Table~\ref{tab:training_settings} summarizes the stage-specific hyperparameters.
Checkpoints are selected by validation pose loss, evaluated on recorded pairs for Stage~1 and rollout pairs for Stages~2 and~3.

% \begin{table}[t]
%     \centering
%     \caption{Training hyperparameters.\todo{need to be improved to make easier to follow, delete}}
%     \label{tab:training_settings}
%     \footnotesize
%     \setlength{\tabcolsep}{2pt}
%     \renewcommand{\arraystretch}{1.2}

%     \begin{tabular*}{\columnwidth}
%         {@{\extracolsep{\fill}}lccc@{}}
%         \hline
%         Pose estimator & Public S1 & Public S2 & In-house S2 \\
%         \hline
%         Max. epochs & 50 & 30 & 50 \\
%         LR: fusion/relation
%         & $5{\times}10^{-4}$ & $10^{-5}$ & $10^{-5}$ \\
%         LR: pose heads
%         & $5{\times}10^{-4}$ & $3{\times}10^{-5}$
%         & $3{\times}10^{-5}$ \\
%         LR: LoRA
%         & $3{\times}10^{-4}$ & $3{\times}10^{-6}$
%         & $3{\times}10^{-6}$ \\
%         \hline
%     \end{tabular*}

%     \smallskip
%     \begin{tabular*}{\columnwidth}
%         {@{\extracolsep{\fill}}lc@{}}
%         \hline
%         Residual head & Public/In-house S3 \\
%         \hline
%         Max. epochs & 20 \\
%         LR: residual head & $3{\times}10^{-4}$ \\
%         % Dropout & 0.10 \\
%         % Weight decay & $10^{-3}$ \\
%         % LR schedule & Cosine; 1 warm-up epoch \\
%         % Early-stopping patience & 10 epochs \\
%         Translation / rotation bounds & 3~mm / $3^\circ$ \\
%         \hline
%     \end{tabular*}
% \end{table}

\begin{table*}[t]
    \centering
    \caption{Overall and distance-stratified reaching success.}
    \label{tab:servo_success_bins}
    \small
    \renewcommand{\arraystretch}{1.12}
    \setlength{\tabcolsep}{4pt}
    \begin{tabular*}{\textwidth}{@{\extracolsep{\fill}}llccccc@{}}
        \hline
        Dataset & Model & Overall $\uparrow$ & 2-4~mm $\uparrow$ & 4-7~mm $\uparrow$ & 7-10~mm $\uparrow$ & 10-15~mm $\uparrow$ \\
        \hline
        Public & A: Initial estimator & 512/720 (71.11\%) & 86.67\% & 71.67\% & 65.00\% & 61.11\% \\
        & B: A + DAgger & 589/720 (81.81\%) & 92.22\% & 80.00\% & 78.89\% & 76.11\% \\
        & C: A + Residual & 594/720 (82.50\%) & 90.56\% & 81.67\% & 82.22\% & 75.56\% \\
        & D: A + DAgger + Residual & \textbf{640/720 (88.89\%)} & \textbf{95.56\%} & \textbf{88.89\%} & \textbf{88.89\%} & \textbf{82.22\%} \\
        \hline
        In-house & Public A (zero-shot) & 189/576 (32.81\%) & 52.1\% & 38.2\% & 27.1\% & 13.9\% \\
        & Public B (zero-shot) & 228/576 (39.58\%) & 59.03\% & 44.44\% & 34.03\% & 20.83\% \\
        & In-house DAgger & 418/576 (72.57\%) & 80.56\% & 74.31\% & 70.14\% & \textbf{65.28\%} \\
        & In-house DAgger + Residual & \textbf{437/576 (75.87\%)} & \textbf{83.33\%} & \textbf{78.47\%} & \textbf{76.39\%} & \textbf{65.28\%} \\
        \hline
    \end{tabular*}
\end{table*}

\begin{table*}[t]
    \centering
    \caption{Failure counts and percentages by termination category.}
    \label{tab:servo_termination}
    \small
    \renewcommand{\arraystretch}{1.12}
    \setlength{\tabcolsep}{4pt}
    \begin{tabular*}{\textwidth}{@{\extracolsep{\fill}}llccc@{}}
        \hline
        Dataset & Model & Out of range $\downarrow$ & Step limit $\downarrow$ & Out of volume $\downarrow$ \\
        \hline
        Public & A: Initial estimator & 16 (2.22\%) & 192 (26.67\%) & \textbf{0 (0.00\%)} \\
        & B: A + DAgger & \textbf{14 (1.94\%)} & 116 (16.11\%) & 1 (0.14\%) \\
        & C: A + Residual & 65 (9.03\%) & 61 (8.47\%) & \textbf{0 (0.00\%)} \\
        & D: A + DAgger + Residual & 30 (4.17\%) & \textbf{50 (6.94\%)} & \textbf{0 (0.00\%)} \\
        \hline
        In-house & Public A (zero-shot) & 170 (29.51\%) & 199 (34.55\%) & 18 (3.13\%) \\
        & Public B (zero-shot) & 154 (26.74\%) & 177 (30.73\%) & 17 (2.95\%) \\
        & In-house DAgger & \textbf{55 (9.55\%)} & 98 (17.01\%) & \textbf{5 (0.87\%)} \\
        & In-house DAgger + Residual & 56 (9.72\%) & \textbf{78 (13.54\%)} & \textbf{5 (0.87\%)} \\
        \hline
    \end{tabular*}
\end{table*}

\begin{table*}[t]
    \centering
    \caption{Control steps to success.}
    \label{tab:servo_steps}
    \small
    \renewcommand{\arraystretch}{1.12}
    \setlength{\tabcolsep}{4pt}
    \begin{tabular*}{\textwidth}{@{\extracolsep{\fill}}llccccc@{}}
        \hline
        Dataset & Model & Overall $\downarrow$ & 2-4~mm $\downarrow$ & 4-7~mm $\downarrow$ & 7-10~mm $\downarrow$ & 10-15~mm $\downarrow$ \\
        \hline
        Public & A: Initial estimator & $6.2\pm9.0$ & $3.8\pm7.1$ & $6.9\pm9.4$ & $6.4\pm8.3$ & $8.7\pm10.7$ \\
        & B: A + DAgger & $5.2\pm6.9$ & $2.4\pm4.3$ & $5.3\pm6.7$ & $5.3\pm5.1$ & $8.3\pm9.6$ \\
        & C: A + Residual & $3.2\pm3.0$ & $1.4\pm1.1$ & $3.5\pm4.1$ & $3.7\pm2.8$ & $4.4\pm2.3$ \\
        & D: A + DAgger + Residual & $\boldsymbol{2.8\pm2.1}$ & $\boldsymbol{1.3\pm1.1}$ & $\boldsymbol{2.6\pm1.7}$ & $\boldsymbol{3.4\pm1.9}$ & $\boldsymbol{4.2\pm2.4}$ \\
        \hline
        In-house & Public A (zero-shot) & 13.7$\pm$12.2 & 8.1$\pm$8.5 & 15.4$\pm$11.1 & 17.9$\pm$13.9 & 22.1$\pm$14.8 \\
        & Public B (zero-shot) & $13.0\pm11.4$ & $7.9\pm8.0$ & $14.1\pm11.7$ & $17.7\pm11.9$ & $17.5\pm12.4$ \\
        & In-house DAgger & $4.8\pm6.6$ & $3.1\pm5.6$ & $4.3\pm7.0$ & $5.8\pm6.6$ & $6.6\pm6.8$ \\
        & In-house DAgger + Residual & $\boldsymbol{3.1\pm3.3}$ & $\boldsymbol{1.6\pm1.4}$ & $\boldsymbol{2.7\pm3.1}$ & $\boldsymbol{3.7\pm3.2}$ & $\boldsymbol{4.9\pm4.1}$ \\
        \hline
    \end{tabular*}
\end{table*}

\subsection{Servoing Results}
\label{sec:virtual_servoing_results}

\subsubsection{Evaluation protocol}
We evaluated target-view reaching in reconstructed volumes from held-out subjects, with initial translation stratified into four distance bins: 2-4~mm, 4-7~mm, 7-10~mm, and 10-15~mm. The public benchmark comprised 720 episodes from five subjects, using 12 scans per subject, three targets per scan, and one initial pose per target and distance bin. The in-house benchmark comprised 576 episodes from two subjects, using the same numbers of scans and targets but two initial poses per bin. Within each dataset, all methods used identical target and initial poses, translation and rotation gains of 0.5, and a 50-step limit. An episode succeeded upon first reaching both a translation error below 2~mm and a rotation error below $3^\circ$.

\par
Table~\ref{tab:servo_success_bins} reports overall and distance-stratified success, with 180 public or 144 in-house episodes per bin. Table~\ref{tab:servo_termination} distinguishes three failure categories: \emph{Out of range}, when translation error exceeds 15~mm; \emph{Step limit}, when the target is not reached within 50 steps; and \emph{Out of volume}, when the sampled slice extends beyond the reconstruction and image coverage falls below 90\%. Failure percentages are calculated over all episodes. Table~\ref{tab:servo_steps} reports steps to success as mean $\pm$ standard deviation over successful episodes only. Arrows indicate the preferred direction, and bold marks the best value within each dataset and column.

\subsubsection{Performance on Public dataset}
The public benchmark isolates the contributions of DAgger adaptation and residual correction through four variants: A, the initial estimator; B, A + DAgger; C, A + Residual; and D, A + DAgger + Residual. Variants B-D use the same collected rollout pool, and the base estimators are frozen during residual training in C and D.

Both components improve reaching success, with the strongest performance obtained by combining them. Relative to A (71.11\%), success increases to 81.81\% with DAgger and 82.50\% with residual correction, while their combination reaches 88.89\%. D performs best across all four distance bins and reduces the mean successful-episode length from 6.2 steps for A to 2.8. These results support complementary contributions to reaching success, accompanied by shorter successful trajectories.

The failure breakdown reveals why adaptation before residual correction is beneficial. Applying residual correction directly to A reduces step-limit failures but increases out-of-range failures. With prior DAgger adaptation, out-of-range failures decrease from 65 for C to 30 for D, while step-limit failures also decrease from 61 to 50. However, B retains the fewest out-of-range failures (14), so the highest reaching success does not coincide with the lowest rate of leaving the permitted motion range.

\subsubsection{Performance on in-house dataset}
% \textit{In-house dataset.}
The in-house benchmark tests transfer from an Ultrasonix system with a curvilinear probe to a SonoScape E2 system with a linear probe, using calibrated $30\times30$~mm ROIs for both datasets. Without in-house training, Public B achieves higher success than Public A (39.58\% versus 32.81\%) and improves performance in every distance bin. This suggests that public-domain DAgger adaptation provides some transferable benefit. Nevertheless, success declines markedly with initial distance, reaching only 13.9\% for A and 20.83\% for B at 10-15~mm, indicating limited zero-shot reaching capability.

Adaptation using in-house scans provides the largest improvement. Starting from Public B, one DAgger round raises success from 39.58\% to 72.57\%, with gains across all distance bins. The largest gain occurs at 10-15~mm, where success reaches 65.28\%. This improvement is accompanied by fewer out-of-range failures (154 to 55) and shorter successful trajectories (13.0 to 4.8 steps on average), supporting the value of adapting to states visited in the target-device reconstructions.

Residual correction provides a further, smaller improvement in success while substantially shortening successful trajectories. Success increases to 75.87\%, and the mean successful-episode length falls from 4.8 to 3.1 steps. The largest additional success gain occurs at 7-10~mm (70.14\% to 76.39\%), whereas the 10-15~mm result remains unchanged. Step-limit failures decrease from 98 to 78, with little change in out-of-range failures (55 versus 56). Thus, in-house adaptation accounts for most of the transfer improvement, while residual correction adds reaching accuracy at intermediate distances and reduces the steps required in successful episodes.

\section{CONCLUSION}
In this study, we proposed \textbf{Recon2Servo}, a novel framework for 6-DoF robotic ultrasound visual servoing that learns image-to-motion inference directly from B-mode images. Inspired by trackerless 3D ultrasound reconstruction, Recon2Servo establishes a spatial mapping between arbitrary current and target ultrasound views without relying on sophisticated handcrafted visual features or anatomy-specific priors. Comprehensive evaluations on the public TUS-REC2024 dataset and an in-house dataset collected from 12 healthy volunteers demonstrated its effectiveness across diverse imaging settings. In unseen virtual servoing experiments on the public dataset, Recon2Servo improved the success rate from 71.11\% to 88.89\%. When transferred to the in-house dataset, the public-data-adapted estimator achieved a zero-shot success rate of 39.58\%, which increased substantially to 75.87\% after closed-loop adaptation and residual correction. Additional real-robot demonstrations of target-view alignment and dynamic tracking on a human forearm are provided in the supplementary video.

% Future work will investigate repeated data aggregation and confidence-aware motion updates to reduce out-of-range failures, and evaluate transfer across more devices and anatomies.


%%%%%%%%%%%%%%%%%%%%%%%%%%%%%%%%%%%%%%%%%%%%%%%%%%%%%%%%%%%%%%%%%%%%%%%%%%%%%%%%

%%%%%%%%%%%%%%%%%%%%%%%%%%%%%%%%%%%%%%%%%%%%%%%%%%%%%%%%%%%%%%%%%%%%%%%%%%%%%%%%

%%%%%%%%%%%%%%%%%%%%%%%%%%%%%%%%%%%%%%%%%%%%%%%%%%%%%%%%%%%%%%%%%%%%%%%%%%%%%%%%



% \section*{ACKNOWLEDGMENT}

% The preferred spelling of the word "acknowledgment" in America is without an "e" after the "g". Avoid the stilted expression, "One of us (R. B. G.) thanks . . ."  Instead, try "R. B. G. thanks". Put sponsor acknowledgments in the unnumbered footnote on the first page.

% Uncomment after adding at least one \cite{...} command.
\bibliographystyle{IEEEtran}
\bibliography{IEEE}

\end{document}